\documentclass{article}

\usepackage{arxiv}

\usepackage[utf8]{inputenc} 
\usepackage[T1]{fontenc}    
\usepackage{hyperref}       
\usepackage{url}            
\usepackage{booktabs}       
\usepackage{amsfonts}       
\usepackage{nicefrac}       
\usepackage{microtype}      
\usepackage{lipsum}		
\usepackage{graphicx}
\usepackage{natbib}
\usepackage{doi}

\usepackage{times}
\usepackage{soul}
\usepackage[small]{caption}
\usepackage{amsmath}
\usepackage{amsthm}
\usepackage{booktabs}
\usepackage{algorithm}
\usepackage{multirow}
\usepackage{makecell}
\usepackage{enumerate} 

\usepackage{subcaption}

\usepackage{float}
\usepackage{xcolor}
\usepackage[export]{adjustbox}

\usepackage{algorithmicx} 
\usepackage{amssymb}
\usepackage[noend]{algpseudocode}
\MakeRobust{\Call}

\title{Explainable Representations for Relation Prediction in Knowledge Graphs}

\author{\href{https://orcid.org/0000-0002-7241-8970}{\includegraphics[scale=0.06]{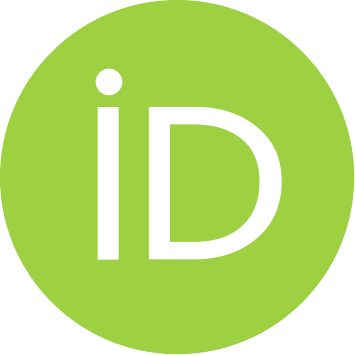}\hspace{1mm}Rita T.Sousa} \\
	LASIGE, Faculdade de Ciências\\
	Universidade de Lisboa\\
	Portugal \\
	\texttt{risousa@ciencias.ulisboa.pt} \\
	\And
	\href{https://orcid.org/0000-0001-8223-4799}{\includegraphics[scale=0.06]{orcid.pdf}\hspace{1mm}Sara Silva} \\
	LASIGE, Faculdade de Ciências\\
	Universidade de Lisboa\\
	Portugal \\
	\texttt{sgsilva@ciencias.ulisboa.pt} \\
        \And
	\href{https://orcid.org/0000-0002-1847-9393}{\includegraphics[scale=0.06]{orcid.pdf}\hspace{1mm}Catia Pesquita} \\
	LASIGE, Faculdade de Ciências\\
	Universidade de Lisboa\\
	Portugal \\
	\texttt{clpesquita@ciencias.ulisboa.pt} \\
}

\renewcommand{\shorttitle}{\textit{arXiv} Template}

\hypersetup{
pdftitle={A template for the arxiv style},
pdfsubject={q-bio.NC, q-bio.QM},
pdfauthor={David S.~Hippocampus, Elias D.~Striatum},
pdfkeywords={First keyword, Second keyword, More},
}

\begin{document}
\maketitle

\begin{abstract}
Knowledge graphs represent real-world entities and their relations in a semantically-rich structure supported by ontologies. Exploring this data with machine learning methods often relies on knowledge graph embeddings, which produce latent representations of entities that preserve structural and local graph neighbourhood properties, but sacrifice explainability. However, in tasks such as link or relation prediction, understanding which specific features better explain a relation is crucial to support complex or critical applications.

We propose SEEK, a novel approach for explainable representations to support relation prediction in knowledge graphs. It is based on identifying relevant shared semantic aspects (i.e., subgraphs) between entities and learning representations for each subgraph, producing a multi-faceted and explainable representation.

We evaluate SEEK on two real-world highly complex relation prediction tasks: protein-protein interaction prediction and gene-disease association prediction.
Our extensive analysis using established benchmarks demonstrates that SEEK achieves significantly better performance than standard learning representation methods while identifying both sufficient and necessary explanations based on shared semantic aspects.
\end{abstract}


\section{Introduction}

Knowledge Graphs (KGs)~\cite{ehrlinger2016towards} are a representation of factual information about entities in the real world and how they relate to each other, having been widely used to support various applications including machine learning~(ML)~\cite{hogan2021knowledge}. Particularly, in scientific domains, KGs have become highly relevant because they allow for the description and linking of information about entities based on ontologies~\cite{staab2010handbook}, allowing the description of complex natural phenomena that are not easily captured in mathematical form~\cite{nicholson2020constructing}.  

In recent years, KG embedding methods~\cite{wang2017knowledge} have become increasingly popular to bridge the gap between the complex representations a KG affords and the vectorial representations most ML methods take as input, since they map KGs into low-dimensional spaces preserving syntactic and structural properties. KG embeddings are popularly employed in link prediction via a scoring function or as features for supervised learning~\cite{portisch2022knowledge}.
However, this represents a significant trade-off: KG embeddings sacrifice the full and rich interpretability offered by KGs, especially when structured by rich ontologies, for the more simple to process latent representations~\cite{palmonari2020knowledge}. 
The effectiveness and usefulness of KG embeddings approach hinges on the crucial assumption that KG embeddings serve as semantically meaningful representations of the underlying entities. To validate such an assumption, KG embedding methods would need to be explainable (i.e., they would need to afford a human-understandable description of the logic, behavior or factors that influence the representation learning process), but in the vast majority of cases they are not. This is a fundamental requirement to ensure the scientific validity of KG embeddings, or any artificial intelligence (AI) method, as a tool that can be used to uncover new knowledge, help understand the mechanisms underlying natural phenomena, and distinguish meaningful predictions from spurious correlations~\cite{barredo2020explainable}.

In this work, we focus specifically on the problem of predicting a relation between KG entities that is not defined in the KG. Predicting relations such as protein-protein interactions (PPI) or gene-disease associations (GDA) by exploring KGs and ontologies has been the focus of extensive research in the biomedical domain. Both algorithmic~\cite{zhang2016,hoehndorf2011phenomenet,asif2018identifying} and ML approaches~\cite{kulmanov2021semantic} have been employed to achieve this with success, with KG embeddings particularly excelling at the task \cite{chen2019protein,teremie2022TransformerGO,alshahrani2017neuro}.
However, understanding the nature of these relations requires discerning which aspects of the KG have the most influence on a prediction.
This empowers users not only in assessing the reliability of the model itself but also in potentially elucidating the phenomena underlying the relation. 
For example, if we were to explain the interaction between the proteins \textit{Protransforming growth factor $\alpha$} and \textit{Disks large homolog 2}, generating an explanation based on the fact that they both perform the very specific function \textit{MAPK cascade}, we would likely increase trust as well as highlight a relevant aspect for interaction. In contrast, a very general explanation, such as the fact that both proteins are present in the \textit{plasma membrane} would contribute to neither.

We propose SEEK (Shared Explainable Embeddings for Knowledge graphs), a novel method for generating explainable KG embeddings that represent entity pairs for relation prediction.
The intuition behind this is that an entity pair can be represented by combining embeddings that represent each of their shared semantic aspects, rather than simply combining their respective embeddings. This technique explores the rich semantics of the ontology to identify the shared semantic aspects between related entities based on computing their disjoint common ancestors. Then, these pair embeddings are used to train a supervised ML model for relation prediction. 
SEEK is fundamentally different from link prediction methods since it produces representations of pairs of entities based on shared semantic aspects, whereas link prediction methods rely on learning representations of individual KG entities and apply a scoring function to estimate the likelihood of triples.

Given a prediction, our method explains it by computing the importance of each shared semantic aspect in identifying it. Inspired by \cite{watson2021local,rossi2022explaining} we consider that an explanation includes two complementary views: the set of semantic aspects that, if absent from an entity pair, would render the model incapable of generating that prediction (i.e., necessary explanations); the set of semantic aspects that, if shared by any entity pair, would prompt the model to produce that prediction (sufficient explanations). Since SEEK explains specific predictions rather than the global mechanism of the model, it consequently falls under the category of local post-hoc explanation methods as proposed by \cite{guidotti2018survey}. 

We evaluate the effectiveness of SEEK in two different tasks, PPI prediction and GDA prediction. Predicting PPI is a crucial task in molecular biology~\cite{chen2021network,hu2021survey}, and several KG embedding-based methods have been employed to tackle it~\cite{kulmanov2019embeddings,smaili2019opa2vec,kulmanov2021semantic,kulmanov2019embeddings,xiong2022faithful}. 
Due to the high costs and challenges involved in experimentally determining PPI, computational methods can be used to identify protein pairs that are likely to interact, which are subsequently validated through experimental assays rendering the process more efficient. Likewise, predicting the relation between genes and diseases is essential to understand disease mechanisms and identifying potential biomarkers or therapeutic targets~\cite{eilbeck2017settling}. Once again, computational approaches to identify the most promising associations to be further validated are commonly employed, with recent approaches applying KG embedding methods~\cite{alshahrani2017neuro,smaili2019opa2vec,nunes2021predicting}. However, opaque methods such as KG embeddings are unable to provide explanations behind each prediction. Explanatory mechanisms can elucidate the potential mechanisms behind the predicted relation, which can be helpful to determine the type of experimental procedure that should be applied to confirm the precited relation but also to identify data biases that can result in misclassification and should be grounds to discard the candidate pair. Our extensive experiments show that our method produces useful explanations besides improving performance over state-of-the-art embedding methods.

Our main contributions are the following:

\begin{itemize}
\item We propose SEEK, a novel method for generating explainable KG embeddings that represent entity pairs for relation prediction.
\item We develop extensions of popular KG embedding methods implementing SEEK.
\item We design explanation methods that quantify the importance of specific KG semantic aspects for specific relation predictions.
\item We report extensive experimental results demonstrating that SEEK is able to produce effective explanations for relation prediction as well as generally improving predictive performance on multiple models and biomedical datasets.
\end{itemize}

\section{Problem Overview}

We define a KG as a labeled directed graph $KG = (V, E, R)$  where $V$ is the set of vertices that represent entities, $R$ is the set of relations and $E$ is the set of edges that connect vertices through relations. Our particular focus is on ontology-rich KGs with ontologies defined using Web Ontology Language (OWL)~\cite{grau2008owl} since biomedical ontologies are typically developed in OWL or have an OWL version. These are frequently found in scientific fields like biomedicine. In these KGs, ontologies are typically used to describe individual instances, while the instances themselves are usually flat with no connections between them. Consequently, there will be two types of vertices in the KG: those that correspond to individual entities and those that correspond to ontology classes, as well as two types of edges: those that relate ontology classes to each other, and those that link individuals to the classes that describe them.  For example, using OWL 2 whose constructs correspond to SROIQ(D), we can indicate that a protein $P$ carries out a function $F$ described in the Gene Ontology (GO) by declaring the axiom $P \sqsubseteq \exists hasFunction.F$. KG embedding methods are then able to learn representations of biomedical entities by exploring the links that connect an entity to the ontology classes that describe it, as well as the structure of the ontology itself.

Our objective is to learn a relation between two KG entities, a pair, when the relation itself is not explicitly defined in the KG, using embeddings as inputs for a supervised ML algorithm. This is a fundamentally distinct task from link prediction, where the training set relations are part of the KG. To tackle this relation prediction task, common approaches typically employ three steps: (1) generate embeddings for each entity in the KG; (2) aggregate the embeddings of each entity in a pair into a single representation; (3) use these aggregated representations as input to a supervised learning algorithm to learn a relation prediction model~\cite{sousa2021evokgsim+,celebi2019evaluation}. 
This generates non-explainable predictions since KG embeddings are, of course, non-explainable, as each dimension does not represent any particular meaning, which poses a serious limitation to the use of KG embeddings in a scientific setting.

Moreover, this particular formulation results in two oversimplifications, which may limit its effectiveness and usefulness. Firstly, it relies on the aggregation of individual embeddings to represent a pair of entities, instead of directly learning an embedding that represents the pair. One should clarify that simply representing the pair as yet another entity on the KG would not be a viable solution, as it would limit the applicability of the approach to pairs seen at representation learning time and thus fail to generalize to novel pairs. Secondly, it focuses on creating an overall representation of each entity, rather than capturing the different semantic aspects that may contribute to the relation we aim to predict. In large KGs, it is not uncommon for entities to be described according to multiple semantic aspects, but only a few may be relevant for the prediction of a particular relation. In a previous study~\cite{sousa2019evolving}, it was demonstrated that not all branches of the GO are equally important for predicting PPIs.

The problem we tackle is then two-fold: (1) to generate latent representations that represent an entity pair directly and (2) to generate latent representations that are amenable to explanation and can capture the relevant semantic aspects for relation prediction.

\section{Related Work}

\subsection{Knowledge Graph Embeddings}

KG embedding methods represent KG entities and their relations in a lower-dimensional space preserving the KG semantic information as much as possible. 
These embeddings have been employed as features in a variety of downstream tasks, such as link prediction, triple classification, or entity typing. 
KG embeddings have been successfully employed in a number of scientific applications, with particular success in the life sciences~\cite{mohamed2021biological,kulmanov2021semantic,chen2019protein,teremie2022TransformerGO}. 
There are several types of KG embeddings, including translational models, semantic matching models, or random walk-based KG embedding approaches. 

Translational methods use distance-based scoring functions. 
TransE~\cite{bordes2013translating} is a well-known approach that assumes the vector of the head entity plus the relation vector should be close to the vector of the tail entity if a relation holds between two entities.
However, TransE only handles one-to-one relationships. To overcome this limitation, TransH~\cite{wang2014knowledge} introduces a unique relation-specific hyperplane for each relationship. 

Semantic matching models rely on scoring functions based on similarity, which can represent the underlying meaning of entities and relationships in vector spaces. An example of such a method is DistMult~\cite{yang2014embedding}, which uses tensor factorization to create vector embeddings for entities and diagonal matrices for relationships.

Random walk-based embedding techniques perform walks in the graph to produce a corpus of sequences that is given as input to a neural language model~\cite{mikolov2013efficient} to learn a latent low-dimensional representation of each entity within the corpus of sequences. RDF2Vec~\cite{ristoski2016rdf2vec} is used to learn embeddings over RDF graphs. 

More recently, KG embedding approaches that tailor representations by considering specific aspects of a KG have been proposed. EL~\cite{kulmanov2019embeddings} and BoxEL~\cite{xiong2022faithful} embeddings are geometric approaches that consider the logical structure of the ontology. OWL2Vec*~\cite{chen2021owl2vec} is very similar to RDF2Vec, but it was designed to learn embeddings of OWL ontologies, which are used to represent knowledge in a more expressive and formal way than RDF graphs. OPA2Vec~\cite{smaili2019opa2vec} considers the lexical portion of the KG, specifically the labels of entities, when generating triples.

\subsection{Explainable Artificial Intelligence Techniques}

The scientific community has long recognized the potential of AI as a tool for scientific discovery, with ML, pattern mining, and reasoning playing crucial roles in several steps of the scientific process~\cite{mjolsness2001machine}. Despite this, the vast majority of scientific projects that utilize AI do not prioritize explainability~\cite{roscher2020explainable}. In the biomedical domain, the complexity of both the data and the natural phenomena under study emphasizes the importance of domain knowledge to support explainability~\cite{holzinger2017we}. A knowledge-enabled explainable AI (XAI) system includes a representation of the domain knowledge specific to the application field. This knowledge is explored for generating explanations that are both comprehensible to users and contextually aware of the mechanistic functioning of the AI system and the knowledge it employs~\cite{chari2020foundations}. 

XAI aims to address several key objectives, including promoting algorithmic fairness, detecting potential biases or issues in training data, ensuring that algorithms function as intended, and bridging the gap between the ML community and other scientific disciplines~\cite{gilpin2018explaining}.
According to~\cite{barredo2020explainable}, XAI approaches can be classified into two types: models that are transparent by design, such as decision trees, linear models, and genetic programming models~\cite{9965435}, or post-hoc explainability techniques that are used to improve the interpretability of models that are not transparent by design. Post-hoc explainability techniques can be categorized as either model-specific or model-agnostic if they are applicable to any ML model. Post-hoc techniques may include visual explanations, explanations by example, explanations by simplification, or feature relevance explanations.

KG embeddings are not explainable, and there is no widely accepted methodology to effectively explain the predictions of KG embeddings~\cite{palmonari2020knowledge}. CRIAGE~\cite{pezeshkpour2019investigating} and Kelpie~\cite{rossi2022kelpie} have made striding efforts towards explaining link prediction based on KG embeddings by identifying the fact to add into or remove from the KG that affects the prediction for a target fact. 
Betz~\textit{et al.}~\cite{betz2022adversarial} also propose a post hoc method that uses adversarial attacks on KG embedding models to identify triples that serve as logical explanations for specific predictions. 
These works differ fundamentally from ours by focusing on single facts about each entity, whereas we focus on shared aspects between entity pairs. Additionally, all of these works face the computational challenge posed by having to retrain the KG model after removing a single fact to explain each prediction, and devise heuristic approaches to minimize this aspect. Our approach does not require retraining the model. Instead, we generate explanations by identifying shared semantic aspects and making predictions with the trained model.
ExCut~\cite{gad2020excut} is another approach that uses KG embeddings to identify clusters of entities and then combines it with rule-mining methods to learn interpretable labels.

\section{Methods}

\subsection{Overview}

SEEK is a novel approach that generates explainable vector representations of KG entity pairs to support relation prediction tasks with minimal loss in performance. Code and online tool are available at \url{https://github.com/liseda-lab/seek}.

\begin{figure}[!ht]
    \centering
    \includegraphics[width=0.6\textwidth]{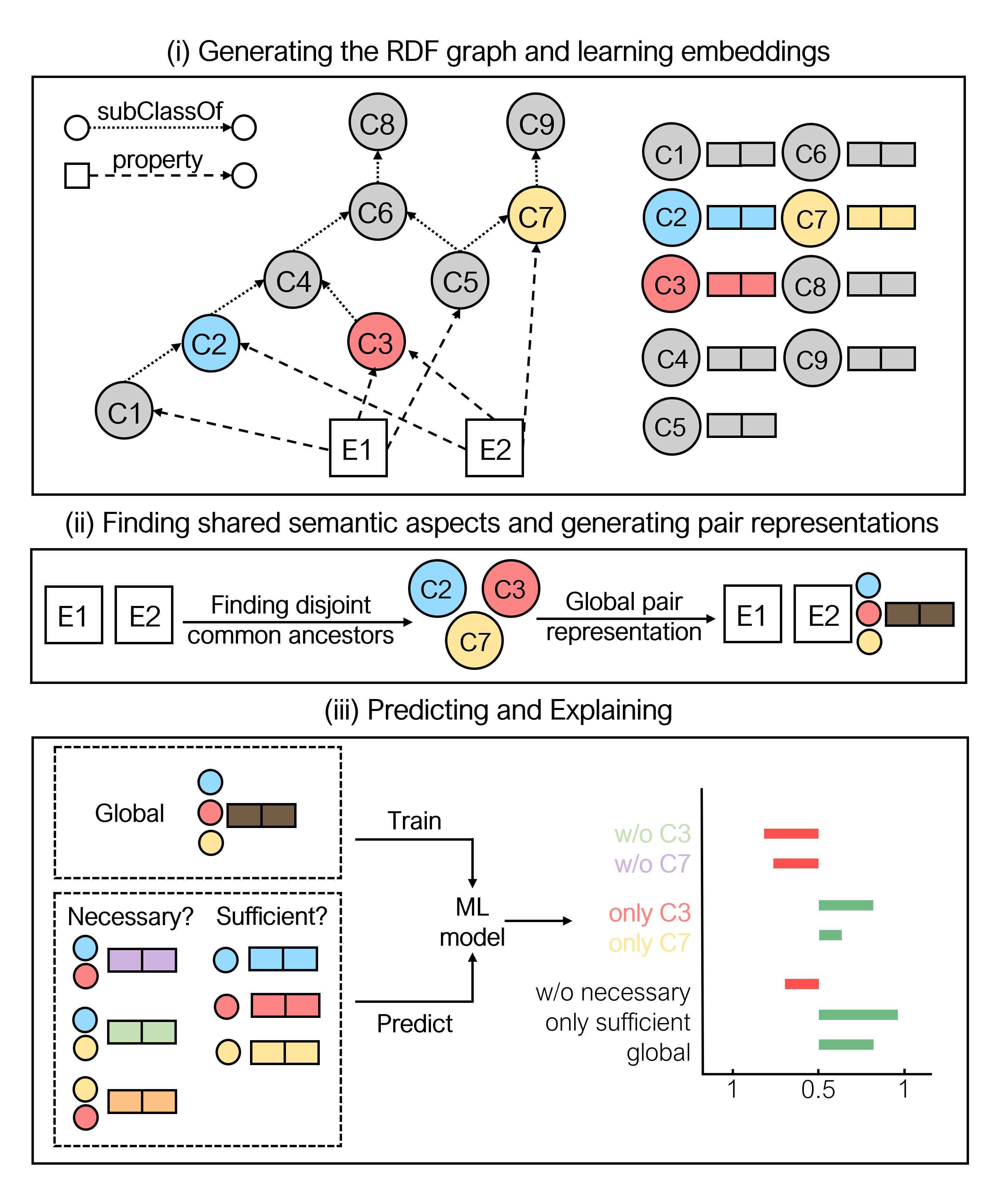}
    \caption{Overview of SEEK with the main steps: (i) generating the RDF graph and learning embeddings (ii) finding shared semantic aspects and generating pair representations (iii) predicting and explaining.}
    \label{fig:methodology}
\end{figure}

Figure~\ref{fig:methodology} shows an overview of the SEEK approach. In the first step, the KG is transformed into an RDF graph, which facilitates the subsequent processing.
Representations for each ontology class are then learned using a KG embedding method. Notably, SEEK is agnostic to the specific KG embedding method employed and can accommodate a broad range of techniques. 

The second step is concerned with identifying the shared semantic aspects between the entities of the pair, which are determined by computing the disjoint common ancestors of all classes related to them. The identification of these semantic aspects is essential for the subsequent generation of accurate and meaningful explanations.
Having identified the relevant semantic aspects, the final representations of entity pairs are then generated by aggregating the embeddings of the shared semantic aspects. 

In the third and final step, supervised learning methods are employed to learn a relation prediction model taking as input the pair embeddings. This model is then used to generate explanations by adopting a perturbation-inspired approach where the contribution of each semantic aspect to the final prediction is assessed in terms of its sufficiency and necessity. The necessary explanations provide insights into the semantic aspects that are necessary for a particular decision to be made, while the sufficient explanations reveal the aspects that are sufficient to support a particular decision. These explanations enable a more thorough understanding of predicted relations and which KG aspects influence it and can be invaluable in identifying potential biases or errors.

\subsection{Generating the RDF Graph and Learning Embeddings}

Ontology-rich KGs are typically defined in OWL. However, the majority of graph processing and analysis tools require RDF graphs. Therefore, the initial step is to convert the KG into an RDF graph following the guidelines provided by the W3C\footnote{https://www.w3.org/TR/owl2-mapping-to-rdf/}.  The conversion process involves transforming simple axioms directly into RDF triples, such as subsumption axioms or data and annotation properties associated with an entity. Multiple triples are created for more complex axioms involving class expressions, which usually require blank nodes.  
The relations between entities and the ontology classes describing them are usually stored in annotation files in the biomedical domain. These annotations are processed into object properties. 
After conversion, the nodes in the RDF graph represent ontology classes or individuals, and the edges represent named relations. Finally, we employ a KG embedding method to learn latent representations of all the ontology classes in the KG. 

\begin{figure}[!ht]
    \centering
    \includegraphics[width=0.4\textwidth]{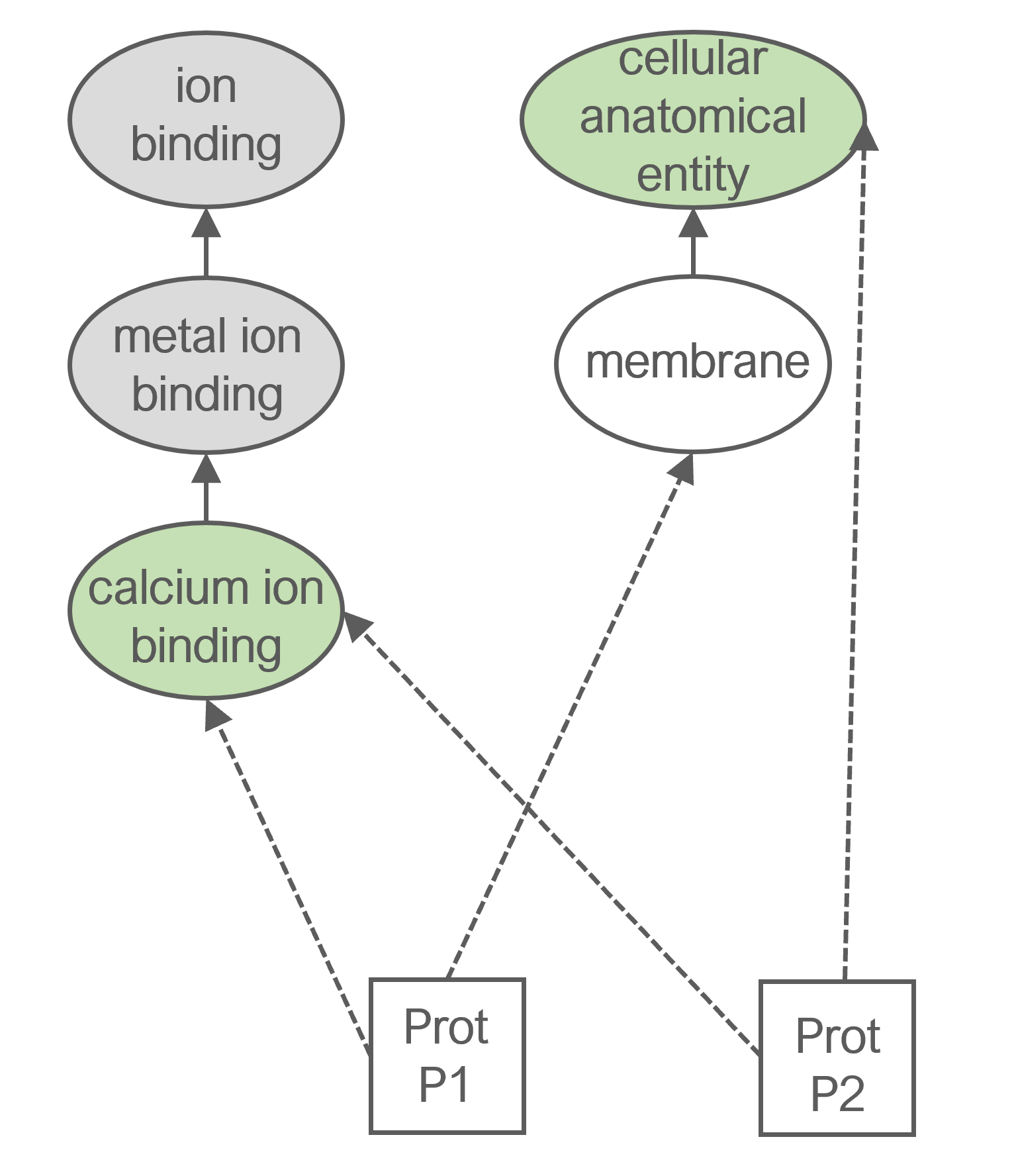}
    \caption{A GO KG subgraph to represent the shared semantic aspects of two entities. Green classes represent the disjoint common ancestors of proteins P1 and P2. Grey classes represent the remaining common ancestors.}
    \label{fig:motivation}
\end{figure}

\subsection{Finding Shared Semantic Aspects and Generating Pair Representations}

To generate a representation for an entity pair we explore the concept of semantic aspect (i.e., a subgraph of the KG that captures a specific perspective of the domain). We propose to represent a pair of KG entities by the set of semantic aspects they share, unfolding their relationship into different dimensions each based on a shared aspect. We define the shared semantic aspects as the set of disjoint common ancestors computed over the set of classes that describe each entity. 

Let us take two entities $e_1$ and $e_2$ and their set of linked classes $C_1$, $C_2$. To compute the set of disjoint shared aspects, we first compute the disjoint common ancestors of $C_1$ and $C_2$. Following~\cite{couto2011disjunctive}, 
we define that $a_1$ and $a_2$ are disjoint common ancestors of a class $c$ if $c \sqsubseteq a_1$, $c \sqsubseteq a_2$, $a_1 \not\sqsubseteq a_2$ and $a_2 \not\sqsubseteq a_1$. We first compute $C_a$, the set of common ancestors between the two sets $C_1$ and $C_2$, and then filter this set to include only the disjoint common ancestors, each of which represents a shared semantic aspect. 
The shared semantic aspects of two sets only include indirect common ancestors if they do not subsume other common ancestors.
Considering the example in Figure~\ref{fig:motivation}, the shared semantic aspects of proteins P1 and P2 correspond to \textit{calcium ion binding} and \textit{cellular anatomical entity}. 

To represent an entity pair we take the embeddings of each class in the shared semantic aspects set and aggregate them using simple operators such as the Hadamard product, the sum, the average or the L1-norm.

\subsection{Predicting and Explaining}

After obtaining the vector representations, we use supervised ML algorithms to learn relation prediction models and ultimately produce explanations for predicted relations. First, we train our model using the global representation of the pair, generated by aggregating all shared semantic aspect embeddings. Then, for each prediction we want to explain, we generate multiple representations that differ by the presence or absence of a semantic aspect. To understand which semantic aspects are necessary for the prediction, we generate representations that remove each aspect in turn (see Algorithm \ref{alg:necessary}), whereas to understand which aspects are sufficient for the prediction, we generate representations that include a single aspect (see Algorithm \ref{alg:sufficient}). A semantic aspect is considered necessary for a prediction if the predicted class changes when it is removed. Likewise, a semantic aspect is considered sufficient for a prediction if the predicted class does not change when it is the only aspect considered.

We define an explanation as the set of the most relevant shared semantic aspects identified as necessary or sufficient.
A necessary explanation is a shared semantic aspect that, when removed from the pair representation, causes the classifier to change its prediction. A sufficient explanation is a shared semantic aspect that, when used alone to represent a pair, causes the classifier to maintain its prediction. A relation may be explained by multiple necessary and sufficient explanations. 

This approach is similar to how saliency XAI methods inject perturbations in the feature space to capture the importance of features. However, it addresses a significant challenge that perturbation or modification-based methods face, including those that aim to explain KG embeddings~\cite{pezeshkpour2019investigating,rossi2022kelpie}, which is the need to relearn representations after performing the modification to the data. SEEK avoids this hurdle since it is based on composite representations of ontology classes, which are easy to modify and do not require retraining since the ontology itself is never altered, so the learned class embeddings remain fixed. 

The final explanation can be represented as a chart where sufficient and necessary shared semantic aspects are presented alongside their impact on the prediction. In Figure~\ref{fig:methodology}, both C3 and C7 are necessary to support the prediction since, without either of them, the prediction value changes when compared to the prediction obtained for the global representation. C3 is also a sufficient aspect since it can single-handedly produce a prediction that agrees with the global one. The explanation can be further enriched with the prediction of the global approach, a prediction made with all sufficient shared semantic aspects, and a prediction made without any of the necessary shared semantic aspects, all predictions including their respective likelihood.

\begin{algorithm}[h!]
\algrenewcommand\algorithmicindent{0.5em}%
\caption{Generation of necessary explanations}
 \textbf{Input:}the entity pair $(e_1, e_2)$;\\
 the KG embedding model $K$;\\
 the relation prediction model $M$;\\
 \textbf{Output:} the set of disjoint shared aspects that are necessary for explaining the prediction
 \begin{algorithmic}[1]
  \State $N \gets empty$ 
  \State $D \gets \Call{get disjoint shared aspects}{(e_1,e_2)}$ 
 \State $E \gets \Call{get embeddings}{K, D}$
 \State $v \gets \Call{aggregate}{E}$
 \State $p \gets \Call{predict}{M,v}$
\For{$d \in D$}
\State $e' \gets E.delete(d)$
 \State $v' \gets \Call{aggregate}{e'}$
 \State $p' \gets \Call{predict}{M,v'}$
 \If{$p \neq p'$}
    \State $N.append(d)$
  \EndIf
 \EndFor
 \Return{$N$}
 \end{algorithmic}
 \label{alg:necessary}
 \end{algorithm}

 \begin{algorithm}[h!]
\algrenewcommand\algorithmicindent{0.5em}%
\caption{Generation of sufficient explanations}
 \textbf{Input:}the entity pair $(e_1, e_2)$;\\
 the KG embedding model $K$;\\
 the relation prediction model $M$;\\
 \textbf{Output:} the set of disjoint shared aspects that are sufficient for explaining the prediction
 \begin{algorithmic}[1]
  \State $S \gets empty$ 
  \State $D \gets \Call{get disjoint shared aspects}{(e_1,e_2)}$ 
 \State $E \gets \Call{get embeddings}{K, d}$
 \State $v \gets \Call{aggregate}{E}$
 \State $p \gets \Call{predict}{M,v}$
\For{$d \in D$}
 \State $v' \gets \Call{get embedding}{K, d}$
 \State $p' \gets \Call{predict}{M,v'}$
 \If{$p == p'$}
    \State $S.append(d)$
  \EndIf
 \EndFor
 \Return{$S$}
 \end{algorithmic}
 \label{alg:sufficient}
 \end{algorithm}

\section{Experimental Results}

\subsection{Experimental Setup}

We evaluate SEEK on two biomedical relation prediction tasks: predicting PPIs and predicting GDA. Both tasks are grounded on ontology-rich KGs, where PPI employs the GO and GDA is based on the Human Phenotype Ontology (HP). 
Additionally, prior studies have shown that different branches of these ontologies have varying impacts on achieving precise predictions~\cite{sousa2019evolving}.

Our work targets relation prediction tasks cast as a classification task that takes as input entity pairs and a KG back-boned by an ontology. Ontologies are arranged in a directed acyclic graph, where ontology classes are connected by subclass relations such that each class is more specific than its ancestor. Moreover, these relationships are transitive, meaning they inherit all ancestors to the root. The data used are described in the following sections.

\begin{table}[!b]
\centering
\setlength{\tabcolsep}{4pt}
\small
\begin{tabular}{lrr}
\toprule
&  \textbf{PPI} &  \textbf{GDA} \\ \cmidrule{2-3}
Ontology classes &  50422 &  15656\\
Literals and blank nodes & 462874  &  443489\\
Instances & 6738 & 4523 \\
Annotations & 349500 & 160009 \\
Positive Pairs & 23571& 8189 \\
Negative Pairs & 23571& 8189 \\
\bottomrule
\end{tabular}
\caption{Statistics for each task regarding classes, nodes, and edges. Positive and negative pairs correspond to the number of positive and negative relations.}
\label{tab:statistics} 
\end{table}

\subsubsection{Protein-Protein Interaction Prediction}

The target relations to predict are obtained from the STRING database~\cite{STRING2021}, one of the largest PPI databases that integrate physical interactions and functional associations between proteins from various sources. We filtered the protein pairs to include only pairs that met the following criteria: (i) each protein must be annotated with the GO, (ii) interactions must be extracted from curated databases or experimentally determined, and (iii) interactions must have a confidence score above 0.950. 
The PPI dataset contains 23571 interacting protein pairs as well as 23571 negative pairs derived from random negative sampling of the same set of proteins.

The GO KG is used to describe proteins and is built by integrating the GO~\cite{GO2021} and protein annotation data~\cite{goa2014}. 
The GO defines a hierarchy of classes that describe protein functions and their relationships.
It can be represented as a graph where nodes are GO classes and edges define relationships between them (e.g.,  $is\_a$; $part\_of$; $has\_part$; $regulates$; $negatively\_regulates$ and $positively\_regulates$), being the majority $is\_a$ relations. 
The three domains of GO (biological processes, molecular functions, and cellular components) are represented as separate root ontology classes since they do not share any common ancestor.
A GO annotation is a statement about the function $F$ of a protein $P$, and it is added in the KG as an assertion ⟨$P, hasFunction, F$⟩. In GO KG, nodes represent proteins or GO classes, and edges represent links between GO classes or annotations. 
Table~\ref{tab:statistics} describes the statistics about PPI data.

\subsubsection{Gene-Disease Association Prediction}

The target relations to predict are obtained from DisGeNET~\cite{pinero2019disgenet}. 
We follow the approach in~\cite{nunes2021predicting}, which excludes associations whose sources are used to create some of the ontology annotations. Moreover, each gene and disease must have at least one HP annotation. 
This resulted in a balanced dataset with a total of 16378 gene-disease pairs.

In this experiment, we employ the HP KG comprising the HP~\cite{HP2021} and HP annotation data to describe genes and diseases.
HP characterizes the phenotypic abnormalities in human hereditary diseases concerning five semantic aspects, namely phenotypic abnormalities, mode of inheritance, clinical course, clinical modifier and frequency. 
Regarding the HP annotations, they link genes and diseases to HP classes and are added in the KG in the same fashion as in the PPI experiment. 
The statistics about GDA data are also shown in Table~\ref{tab:statistics}.

\subsubsection{Models}

SEEK is independent of the KG embedding method and of the supervised ML algorithm.  For our experiments, we implemented five representative KG embeddings covering translational, semantic matching and random walk-based methods:
 RDF2Vec~\cite{ristoski2016rdf2vec}, OWL2Vec*~\cite{chen2021owl2vec}, TransE~\cite{bordes2013translating}, TransH~\cite{wang2014knowledge} and distMult~\cite{yang2014embedding}.
 RDF2Vec and OWL2Vec* are path-based approaches adapted to RDF graphs that employ neural language models over random walks on the graph. TransE and TransH are translational distance embedding approaches that exploit distance-based scoring functions.
distMult is a semantic matching approach that exploits similarity-based scoring functions.

To generate a pair representation, we use the average as the aggregation which ensures that the values of each dimension remain within the distribution. In the case of necessary explanations, removing one similar semantic aspect will result in a very similar aggregated representation, revealing that the semantic brings little novel information for the prediction (since a similar semantic aspect is still considered). In the case of sufficient explanations, semantic aspects are evaluated independently.

As supervised ML algorithms, we employ two ensemble methods,  Random Forest (RF)~\cite{RF2001} and eXtreme Gradient Boosting (XGB)~\cite{XGB2016}, and a neural network-based method, Multilayer Perceptron (MLP)~\cite{MLP1986}.

\subsection{Results and Discussion}
 
\subsubsection{Performance Evaluation}
To assess our method, we compare the relation prediction performance of our pair representations against the state-of-the-art approach of entity vector aggregation using representative KG embedding methods, supervised ML algorithms and the Hadamard operator. We do not compare SEEK to other KG embedding explanation methods such as Kelpie or CRIAGE because they learn embeddings that target link prediction, whereas SEEK learns embeddings to serve as features for supervised ML.
We evaluate the predictive performance of our approach against our baselines for each task using 10-fold cross-validation. For each partition, the precision (Pr), recall (Re) and weighted average f1-score (F1) are computed, and we report the median of the obtained scores (Table~\ref{tab:performance}) and statistical significance of the observed differences.

The results demonstrate that SEEK outperforms the baseline in all cases but one for PPI prediction, while achieving similar or improved scores for GDA. Curiously, the performance of translational methods shows a marked improvement when using SEEK, likely due to the fact that these methods struggle with learning entity representations, but not ontology class representations.

\begin{table*}[!ht]
\small
\centering
\setlength{\tabcolsep}{4pt}
\begin{tabular}{llrrrlrrrlrrrlrrr}

\toprule

                         && \multicolumn{7}{c}{\textbf{PPI Prediction}} && \multicolumn{7}{c}{\textbf{GDA Prediction}} \\\cmidrule{3-9} \cmidrule{11-17}
                          &     & \multicolumn{3}{c}{Baseline} &  & \multicolumn{3}{c}{SEEK} && \multicolumn{3}{c}{Baseline} &  & \multicolumn{3}{c}{SEEK} \\ \cmidrule{3-5} \cmidrule{7-9} \cmidrule{11-13} \cmidrule{15-17}
\multicolumn{1}{l}{}      &     & Pr   & Re  & F1                   &  & Pr     & Re     & F1    &     & Pr   & Re  & F1                   &  & Pr     & Re     & F1                  \\
\multirow{3}{*}{RDF2Vec}  & XGB &   0.905 & 0.917 & 0.910 &  &  \underline{ 0.920} & 0.910       &  \underline{ 0.915} &  & 0.736 & 0.708 & 0.724 &  &  \underline{ 0.772} & 0.626       & 0.719                       \\
                          & RF  &  0.921 & 0.881 & 0.902 &  & 0.922       &  \underline{ 0.892} &  \underline{ 0.910} &  & 0.783 & 0.673 & 0.740 &  & 0.787       & 0.625       & 0.723 \\
                          & MLP &  0.897 & 0.907 & 0.902 &  &  \underline{ 0.908} &  \underline{ 0.924} &  \underline{ 0.917} &  & 0.700 & 0.705 & 0.696 &  &  \underline{ 0.730} & 0.645       & 0.703\\ \midrule
\multirow{3}{*}{OWL2Vec*}  & XGB &   0.890 & 0.881 & 0.888 &  &  \underline{ 0.933} &  \underline{ 0.925} &  \underline{ 0.929} &  & 0.700 & 0.664 & 0.688 &  &  \underline{ 0.780} & 0.647       &  \underline{ 0.728}   \\
                          & RF  &   0.913 & 0.832 & 0.875 &  &  \underline{ 0.922} &  \underline{ 0.915} &  \underline{ 0.919} &  & 0.730 & 0.618 & 0.690 &  &  \underline{ 0.780} &  \underline{ 0.662} &  \underline{ 0.737}\\
                          & MLP &   0.872 & 0.865 & 0.869 &  &  \underline{ 0.934} &  \underline{ 0.923} &  \underline{ 0.931} &  & 0.648 & 0.676 & 0.650 &  &  \underline{ 0.749} & 0.642       &  \underline{ 0.720}\\ \midrule
\multirow{3}{*}{distMult}   & XGB &  0.897 & 0.905 & 0.902 &  &  \underline{ 0.914} &  \underline{ 0.910} &  \underline{ 0.912} &  & 0.718 & 0.668 & 0.704 &  &  \underline{ 0.764} & 0.649       &  \underline{ 0.722} \\
                          & RF  &   0.904 & 0.860 & 0.884 &  &  \underline{ 0.910} &  \underline{ 0.897} &  \underline{ 0.905} &  & 0.745 & 0.636 & 0.706 &  &  \underline{ 0.766} & 0.637       &  \underline{ 0.716}    \\
                          & MLP &   0.894 & 0.894 & 0.896 &  &  \underline{ 0.881} &  \underline{ 0.895} & 0.888       &  & 0.731 & 0.681 & 0.715 &  & 0.768       & 0.589       & 0.698   \\ \midrule
\multirow{3}{*}{TransE}   & XGB &   0.642 & 0.613 & 0.638 &  &  \underline{ 0.914} &  \underline{ 0.912} &  \underline{ 0.913} &  & 0.526 & 0.509 & 0.524 &  &  \underline{ 0.755} &  \underline{ 0.650} &  \underline{ 0.721}  \\
                          & RF  &   0.590 & 0.542 & 0.583 &  &  \underline{ 0.908} &  \underline{ 0.900} &  \underline{ 0.905} &  & 0.505 & 0.474 & 0.502 &  &  \underline{ 0.765} &  \underline{ 0.640} &  \underline{ 0.719} \\
                          & MLP &   0.250 & 0.500 & 0.333 &  &  \underline{ 0.882} & 0.899       &  \underline{ 0.890} &  & 0.500 & 1.000 & 0.333 &  &  \underline{ 0.779} & 0.555       &  \underline{ 0.694}  \\ \midrule
\multirow{3}{*}{TransH} & XGB &   0.642 & 0.614 & 0.637 &  &  \underline{ 0.921} &  \underline{ 0.918} &  \underline{ 0.919} &  & 0.511 & 0.493 & 0.510 &  &  \underline{ 0.767} &  \underline{ 0.651} &  \underline{ 0.726} \\
                          & RF  &  0.586 & 0.551 & 0.579 &  &  \underline{ 0.912} &  \underline{ 0.908} &  \underline{ 0.910} &  & 0.500 & 0.453 & 0.494 &  &  \underline{ 0.770} &  \underline{ 0.642} &  \underline{ 0.720} \\
                          & MLP &   0.250 & 0.500 & 0.333 &  &  \underline{ 0.915} & 0.920       &  \underline{ 0.920} &  & 0.000 & 0.000 & 0.333 &  &  \underline{ 0.735} &  \underline{ 0.665} &  \underline{ 0.711} \\                
\bottomrule
\end{tabular}
\caption{Medians of precision, recall, and weighted average f1-score (Pr, Re, F1) comparing our approach SEEK to the baseline when coupled with different supervised ML methods for PPI and GDA prediction. SEEK performance values are underlined when improvements are statistically significant with $p$-value $< 0.05$ for the Wilcoxon test against the baselines.}
\label{tab:performance}
\end{table*}

\begin{figure}[!ht]
    \centering 
    \begin{subfigure}{0.45\columnwidth}
        \includegraphics[scale=0.25]{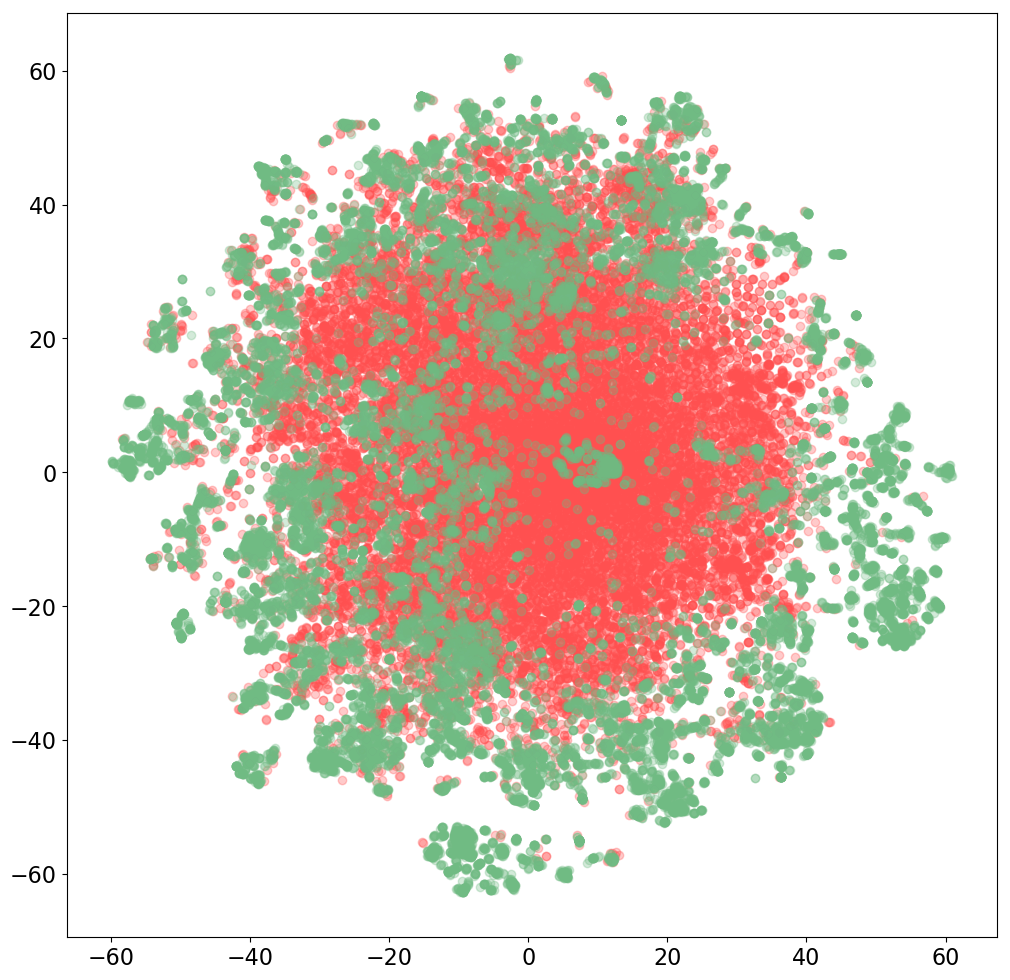}
        \caption{Baseline for PPI}
        \label{fig:baselinePPI}
    \end{subfigure}
\hfill
    \begin{subfigure}{0.45\columnwidth}
        \includegraphics[scale=0.25]{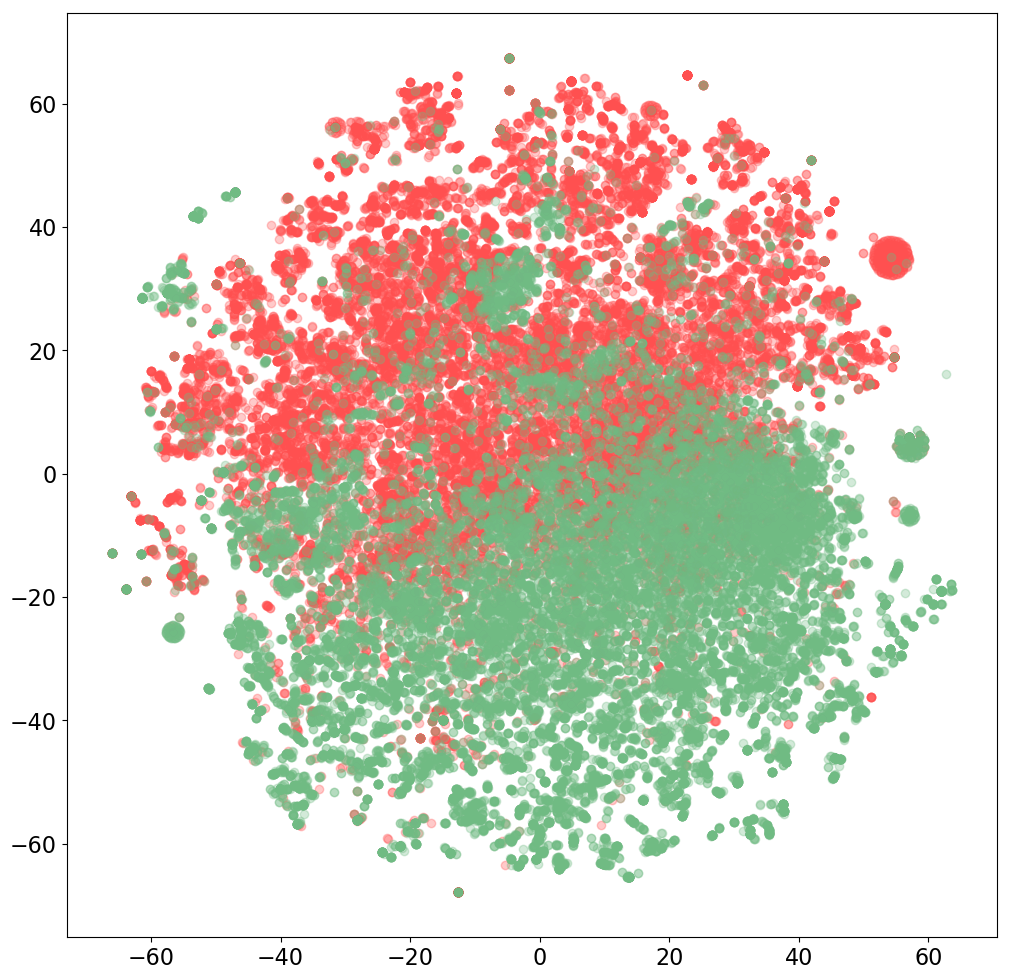}
        \caption{SEEK for PPI}
        \label{fig:approachPPI}
    \end{subfigure}
\hfill
    \begin{subfigure}[b]{0.45\columnwidth}
        \includegraphics[scale=0.25]{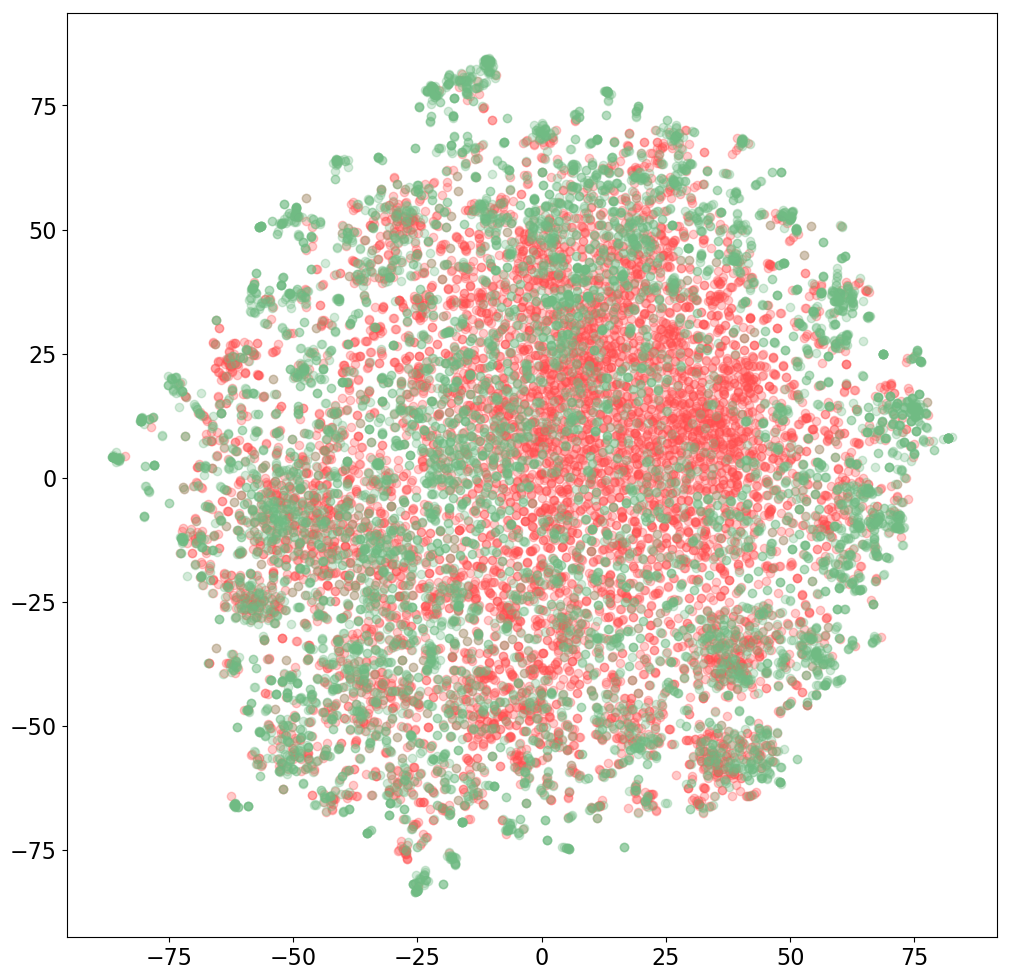}
        \caption{Baseline for GDA}
        \label{fig:baselineGDA}
    \end{subfigure}
 \hfill
    \begin{subfigure}[b]{0.45\columnwidth}
        \includegraphics[scale=0.25]{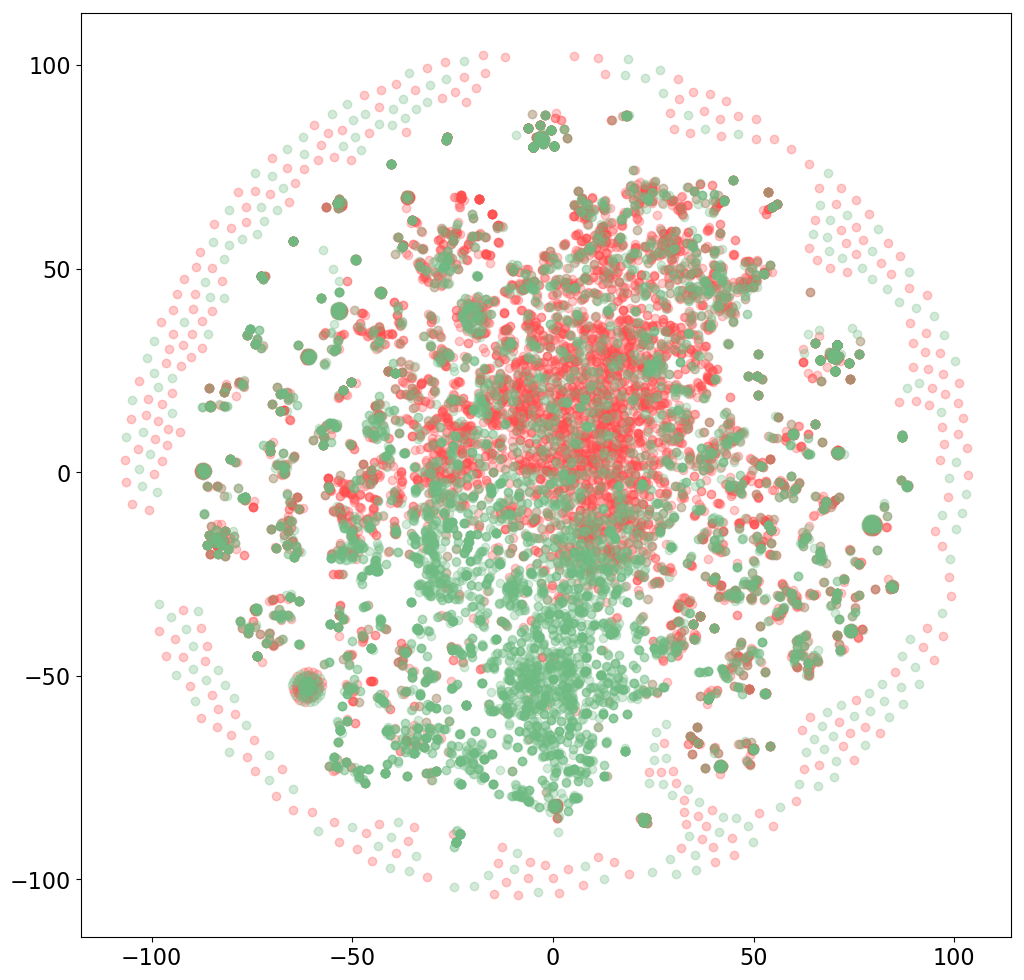}
        \caption{SEEK for GDA}
        \label{fig:approachGDA}
    \end{subfigure}
    \caption{t-SNE plots comparing SEEK to the baseline using RDF2Vec. Positive pairs in green and negative pairs in red.}
    \label{fig:TSNEplots}
\end{figure}

To better understand the differences between the pair representations obtained using the baselines and the ones obtained using SEEK, we plot the RDF2Vec embeddings using \mbox{t-SNE}~\cite{van2008visualizing}, a nonlinear dimensionality reduction technique that is particularly well-suited for visualizing high-dimensional data (Figure~\ref{fig:TSNEplots}). These plots show that our pair representations decrease the overlap between positive and negative pairs and thus are likely to be capturing more meaningful representations.

\subsubsection{Effectiveness of Explanations}

The effectiveness of the explanations is measured based on how predictive performance varies under two scenarios: when pairs are represented without the \textit{necessary} shared semantic aspects; when pairs are represented by \textit{sufficient} shared semantic aspects only. Table~\ref{tab:effectiveness} presents the results obtained for the PPI and GDA tasks using the two best performing KG embedding methods.

In the necessary scenario, we extract the necessary explanations for all correctly predicted relations and produce an ablated representation that does not include any of the necessary shared semantic aspects.  
The performance variation, in terms of precision (Pr), recall (Re) and F1-score (F1), is measured as the difference in predictive performance between the global representation and the ablated representation. The more negative $\Delta$ Pr,  $\Delta$ Re or $\Delta$ F1 are, the more effective are the necessary explanations.

In the sufficient scenario, we extract the sufficient explanations for all incorrectly predicted relations and produce an ablated representation that only includes the sufficient shared semantic aspects.    
The performance variation is also measured as the difference in predictive performance between the global representation and the ablated representation, but in this case the performance of the global representation is actually zero for all scores, since this is only applied to incorrectly predicted relations. A higher $\Delta$ value indicates increased effectiveness.

\begin{table*}[!ht]
\small
\centering
\setlength{\tabcolsep}{4pt}
\begin{tabular}{llrrrlrrrlrrrlrrr}
\toprule
&& \multicolumn{7}{c}{\textbf{PPI Prediction}} && \multicolumn{7}{c}{\textbf{GDA Prediction}} \\\cmidrule{3-9} \cmidrule{11-17}
                                                &           & \multicolumn{3}{c}{RDF2Vec}      &  & \multicolumn{3}{c}{OWL2Vec*}  &           & \multicolumn{3}{c}{RDF2Vec}      &  & \multicolumn{3}{c}{OWL2Vec*}    \\ \cmidrule{3-5} \cmidrule{7-9} \cmidrule{11-13} \cmidrule{15-17}
\multicolumn{1}{l}{}                            &           & MLP & XGB & RF                   &  & MLP & XGB & RF   &           & MLP & XGB & RF                   &  & MLP & XGB & RF                \\
\multirow{3}{*}{\parbox{1.5cm}{w/o \\necessary}}                  & $\Delta$Pr        &    -0.157 & -0.109 & -0.106 &  & -0.095 & -0.099 & -0.089 & &   -0.291 & -0.296 & -0.326 &  & -0.265 & -0.332 & -0.269\\
                                                & $\Delta$Re        &    -0.137 & -0.120 & -0.153 &  & -0.145 & -0.131 & -0.129 & &   -0.329 & -0.220 & -0.277 &  & -0.353 & -0.208 & -0.329\\
                                                & $\Delta$F1       &     -0.148 & -0.113 & -0.125 &  & -0.117 & -0.113 & -0.107 & &    -0.264 & -0.225 & -0.273 &  & -0.270 & -0.256 & -0.260 \\\midrule   
\multirow{3}{*}{\parbox{1.5cm}{only \\sufficient}}                  & $\Delta$Pr        &     0.932  & 1.000  & 0.973  &  & 0.981  & 1.000  & 0.988 & &   0.957  & 0.969  & 0.893  &  & 0.921  & 0.986  & 0.917 \\
                                                & $\Delta$Re        &     0.959  & 1.000  & 0.888  &  & 0.927  & 1.000  & 0.942 & &   0.737  & 0.905  & 0.779  &  & 0.777  & 0.993  & 0.869 \\
                                                & $\Delta$F1       &     0.950  & 1.000  & 0.945  &  & 0.954  & 1.000  & 0.967 & &    0.898  & 0.964  & 0.896  &  & 0.885  & 0.993  & 0.925\\ 
\bottomrule
\end{tabular}
\caption{Explanation efectiveness measured based on the precision (Pr), recall (Re) and weighted average f1-score (F1) variation for PPI and GDA prediction.}
\label{tab:effectiveness}
\end{table*}

\subsubsection{Explanation Length}
The lengths of the explanations, as measured by the number of shared semantic aspects that compose them, are presented in Tables~\ref{tab:PPIexplanationsize} and~\ref{tab:GDAexplanationsize}.  In both tasks, the length of necessary explanations is markedly lower than the length of sufficient explanations, highlighting that for many relations there are no necessary shared semantic aspects. When comparing the shown results to the original number of shared semantic aspects, 9.1 ($\pm$ 6.5) for PPI and 8.5 ($\pm$ 11.0) for GDA, we can verify that sufficient explanations amount to roughly 30\% of shared semantic aspects. These sizes are congruent with the number of objects (7 $\pm$ 2) humans are able to hold in short-term memory according to cognitive studies~\cite{miller1956magical}.

\begin{table}[!ht]
\small
\centering
\setlength{\tabcolsep}{4pt}
\begin{tabular}{llrrlrr}
\toprule
                                                &           & \multicolumn{2}{c}{RDF2Vec}      &  & \multicolumn{2}{c}{OWL2Vec*}      \\ \cmidrule{3-4} \cmidrule{6-7}
\multicolumn{1}{l}{}                            &           & Avg & Std                   &  & Avg & Std                   \\

\multirow{3}{*}{\parbox{1.5cm}{\textbf{sufficient}}}                   & MLP        &    5.6 & 3.9  && 5.3 & 3.5\\
                                                                             & XGB        &    6.2 & 3.9  &&  6.3 & 4.1\\
                                                                             & RF       &      5.6 & 3.7  && 5.9 & 3.7\\\midrule   
\multirow{3}{*}{\parbox{1.5cm}{\textbf{necessary}}}                  & MLP        &     0.4 & 1.1  && 0.3 & 1.0\\
                                                                           & XGB        &    0.4 & 1.1  && 0.3 & 1.0\\
                                                                           & RF       &     0.4 & 1.3  && 0.3 & 1.1\\ 

\bottomrule
\end{tabular}
\caption{Explanation average length (Avg) and standard deviation (Std) for PPI prediction.}
\label{tab:PPIexplanationsize}
\end{table}

\begin{table}[!ht]
\centering
\small
\setlength{\tabcolsep}{4pt}
\begin{tabular}{llrrlrr}
\toprule
                                                &           & \multicolumn{2}{c}{RDF2Vev}      &  & \multicolumn{2}{c}{OWL2Vec*}      \\ \cmidrule{3-4} \cmidrule{6-7}
\multicolumn{1}{l}{}                            &           & Avg & Std                   &  & Avg & Std                   \\ 
\multirow{3}{*}{\parbox{1.5cm}{\textbf{sufficient}}}                   & MLP        &  5.6   &  7.1  && 5.5 &     7.8\\
                                                                             & XGB        &    6.0   &  8.3 && 6.0  &   9.4\\
                                                                             & RF       &    5.6  &  7.7  &&  5.7   &   8.6\\\midrule   
\multirow{3}{*}{\parbox{1.5cm}{\textbf{necessary}}}                  & MLP        &    0.6  & 1.5 && 0.6  &  1.1\\
                                                                           & XGB        &  0.5  & 1.3 && 0.5  &    1.2 \\
                                                                           & RF       &    0.7    &  1.7 && 0.7   &   1.4 \\
\bottomrule
\end{tabular}
\caption{Explanation average length (Avg) and standard deviation (Std) for GDA prediction.}
\label{tab:GDAexplanationsize}
\end{table}

\subsubsection{Examples of Explanations}

Table~\ref{tab:explanationsbyexample} presents explanations for four protein pairs chosen randomly from the PPI dataset. Each pair represents each of the four possible outcomes: a true positive, a false positive, a true negative, and a false negative. 

\begin{table*}[!ht]
\centering
\begin{adjustbox}{max width=\textwidth}

\begin{tabular}{l}

\toprule
       \Large{\textbf{Paxillin -- Integrin $\alpha$-4}}\\ (True positive) \\ \toprule
         \hspace{3.2cm}\includegraphics[width=0.9\linewidth,center]{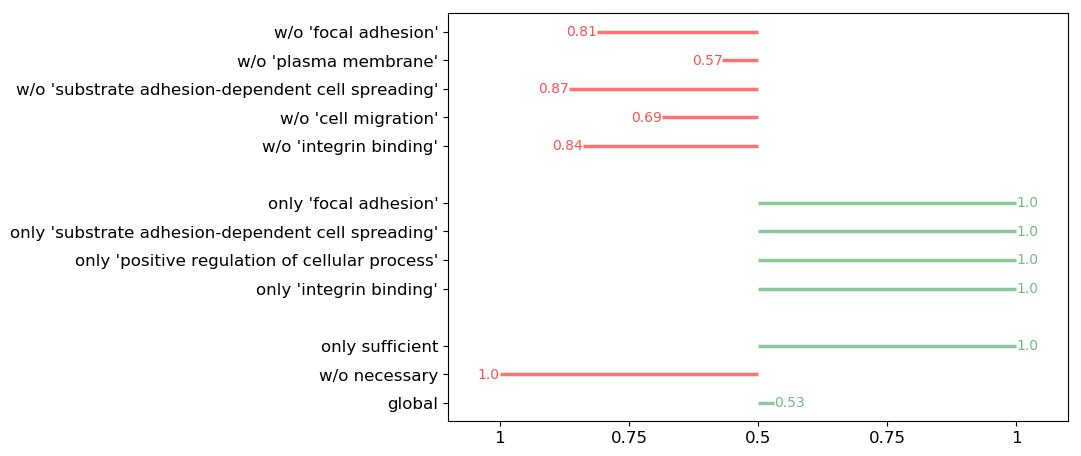}
            \\ \midrule

        \Large{\textbf{Pulmonary surfactant-associated protein B -- Granulocyte-macrophage colony-stimulating factor receptor subunit $\alpha$}} \\(False negative) \\ \toprule
          \hspace{4cm}\includegraphics[width=0.84\linewidth,center]{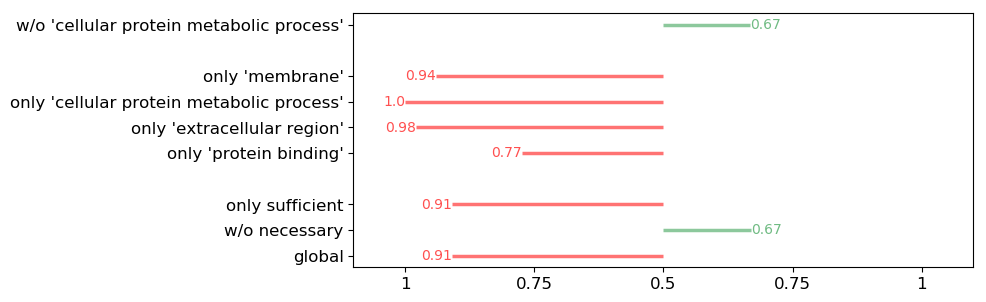}
            \\ \midrule

    \Large{\textbf{Proline-rich 5-like -- Guanine nucleotide-binding 3-like}} \\(True Negative) \\ \toprule
          \hspace{2cm}\includegraphics[width=1.05\linewidth,center]{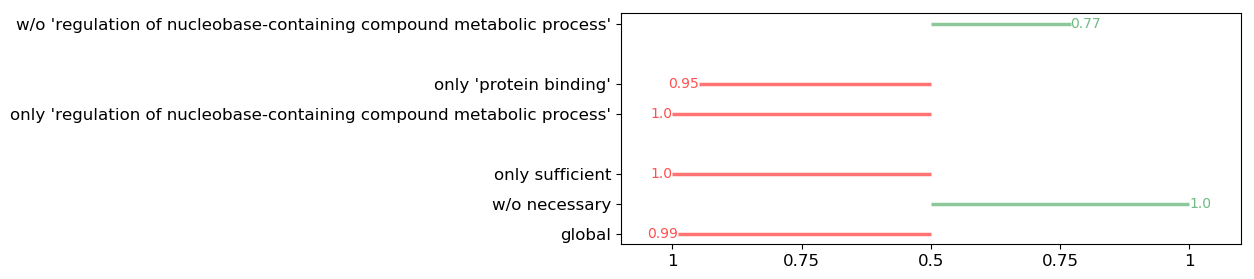}\\
    \midrule

\Large{\textbf{Protransforming growth factor $\alpha$ -- Disks large homolog 2}} \\(False Positive) \\ \toprule
         \hspace{3.5cm}\includegraphics[width=0.89\linewidth,center]{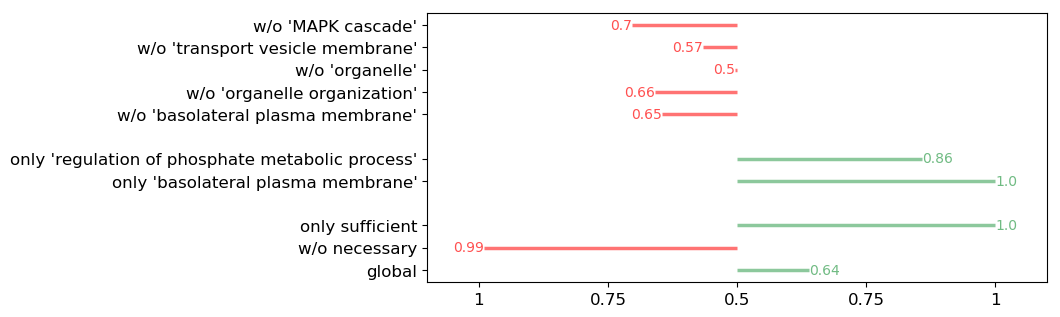}
  \\ 
\bottomrule
\end{tabular}
\end{adjustbox}
\caption{Explanations of PPI prediction models for four randomly selected pairs. For each pair, we provide a bar chart using different sets of disjoint common ancestors to represent the pair. On the x-axis, each bar represents the likelihood returned by the MLP model of the predicted class being correct. Classes are represented by colors (red for class 0 and green for class 1).}
\label{tab:explanationsbyexample}
\end{table*}

The first pair in our analysis consists of paxillin\footnote{https://www.uniprot.org/uniprot/P49023} and integrin $\alpha$-4\footnote{https://www.uniprot.org/uniprot/P13612}.
There is strong evidence for their interaction~\cite{HAN200140903} since integrin $\alpha$-4 binds tightly to paxillin, leading to increased cell migration and an altered cytoskeletal organization that results in reduced cell spreading. Our explanations identify several aspects that are necessary and/or sufficient to explain the interaction and that strongly correlate with the known evidence: focal adhesion, substrate adhesion-dependent cell spreading, cell migration and integrin binding.

The proteins Pulmonary surfactant-associated protein B\footnote{https://www.uniprot.org/uniprot/P07988} and ganulocyte-macrophage colony-stimulating factor receptor subunit $\alpha$\footnote{https://www.uniprot.org/uniprot/P15509} make up the second pair. Although the proteins share some necessary and/or sufficient semantic aspects, they are very general; therefore, the model does not predict the interaction. However, according to the literature, they are likely involved in the same pulmonary disease~\cite{trapnell2003pulmonary}. Both proteins are poorly described under the GO, which can explain why the relation prediction model fails. 

The third pair includes the proline-rich 5-like protein\footnote{https://www.uniprot.org/uniprot/Q6MZQ0} and the guanine nucleotide-binding 3-like protein\footnote{https://www.uniprot.org/uniprot/Q9NVN8}. The model predicts this as a negative pair, and the explanations confirm this, with the removal of the necessary shared semantic aspect resulting in a positive prediction. No interaction is known between these two proteins. 

The Protransforming growth factor (TGF) $\alpha$\footnote{https://www.uniprot.org/uniprot/P01135} and the Disks large homolog  2 (Dlg2)\footnote{https://www.uniprot.org/uniprot/Q15700} compose the last pair and correspond to a false positive. The explanations highlight their participation in the MAPK cascade (central signaling pathways that regulate a wide variety of stimulated cellular processes, including proliferation, differentiation, apoptosis and stress response) as well as their co-location in the basolateral plasma membrane.  It is intriguing to note that although there is no known interaction between these proteins, there is evidence of an interaction between highly similar proteins: TGF-$\beta$ is regulated by Dlg5 and both proteins activate the MAPK cascade~\cite{SEZAKI20131624}. We hypothesize this is not in fact a true negative pair but a still unknown PPI erroneously used as a negative example through random negative sampling.

\section{Conclusion}

Existing KG embedding methods are not explainable, which hinders their application in complex and critical domains. This is especially challenging in relation prediction, where understanding which KG semantic aspects are more relevant for a relationship between two KG entities can provide insightful knowledge about its mechanisms and help distinguish meaningful predictions from spurious correlations. 

To address this challenge, we propose SEEK, a novel approach for learning and explaining representations of KG entity pairs based on their shared semantic space for relation prediction. Its explanatory mechanism is based on generating perturbed representations to identify the relevant semantic aspects of the KG that explain a relation; and since it does not require retraining of representations, it is particularly efficient.
We evaluate SEEK on protein-protein interaction prediction and gene-disease association prediction, two complex and core tasks in the biomedical domain.
SEEK clearly outperforms state-of-the-art learning representation methods in performance, while generating explanations that can identify critical factors driving biological phenomena.

In future work, we will conduct user studies with biomedical domain experts to evaluate SEEK explanations and also improve explanations by investigating the minimal set of shared semantic aspects required to adequately explain a relation.

\section*{Acknowledgments}

C.P., S.S., and R.T.S. are funded by FCT, Portugal, through LASIGE Research Unit (ref. UIDB/00408/2020 and ref. UIDP/00408/2020). R.T.S. acknowledges the FCT PhD grant (ref. SFRH/BD/145377/2019). This work was also partially supported by the KATY project, which has received funding from the European Union’s Horizon 2020 research and innovation programme under grant agreement No 101017453, and by HfPT: Health from Portugal under the Portuguese Plano de Recuperação e Resiliência.

\bibliographystyle{unsrtnat}
\bibliography{references}  






\end{document}